\documentclass[conference]{IEEEtran}

\ifCLASSINFOpdf

\else

\fi

\usepackage{hyperref}
\hypersetup{
	colorlinks = true
}

\usepackage{graphicx}
\usepackage{algorithm}
\usepackage{amssymb}
\usepackage[noend]{algpseudocode}
\usepackage{subfigure}
\usepackage[font=small,labelfont=bf]{caption} 

\begin{document}
%
\title{Rethinking Table Recognition using \\Graph Neural Networks}


\author{\IEEEauthorblockN{Shah Rukh Qasim\IEEEauthorrefmark{1},
Hassan Mahmood\IEEEauthorrefmark{1}, and Faisal Shafait\IEEEauthorrefmark{1},\IEEEauthorrefmark{2}}
\IEEEauthorblockA{
\IEEEauthorrefmark{1}School of Electrical Engineering and Computer Science (SEECS)\\
National University of Sciences and Technology (NUST), Islamabad, Pakistan\\
\IEEEauthorrefmark{2}Deep Learning Laboratory, National Center of Artificial Intelligence (NCAI), Islamabad, Pakistan\\
Email: faisal.shafait@seecs.edu.pk}}



\maketitle

\begin{abstract}
Document structure analysis, such as zone segmentation and table recognition, is a complex problem in document processing and is an active area of research. The recent success of deep learning in solving various computer vision and machine learning problems has not been reflected in document structure analysis since conventional neural networks are not well suited to the input structure of the problem. In this paper, we propose an architecture based on graph networks as a better alternative to standard neural networks for table recognition. We argue that graph networks are a more natural choice for these problems, and explore two gradient-based graph neural networks. Our proposed architecture combines the benefits of convolutional neural networks for visual feature extraction and graph networks for dealing with the problem structure. We empirically demonstrate that our method outperforms the baseline by a significant margin. In addition, we identify the lack of large scale datasets as a major hindrance for deep learning research for structure analysis and present a new large scale synthetic dataset for the problem of table recognition. Finally, we open-source our implementation of dataset generation and the training framework of our graph networks to promote reproducible research in this direction\footnotemark.
\footnotetext{\href{https://github.com/shahrukhqasim/TIES-2.0}{github.com/shahrukhqasim/TIES-2.0}}

\end{abstract}


\IEEEpeerreviewmaketitle

\begin{figure*}[ht]
   \centering
   \subfigure[Category 1]{\includegraphics[width=.35\textwidth]{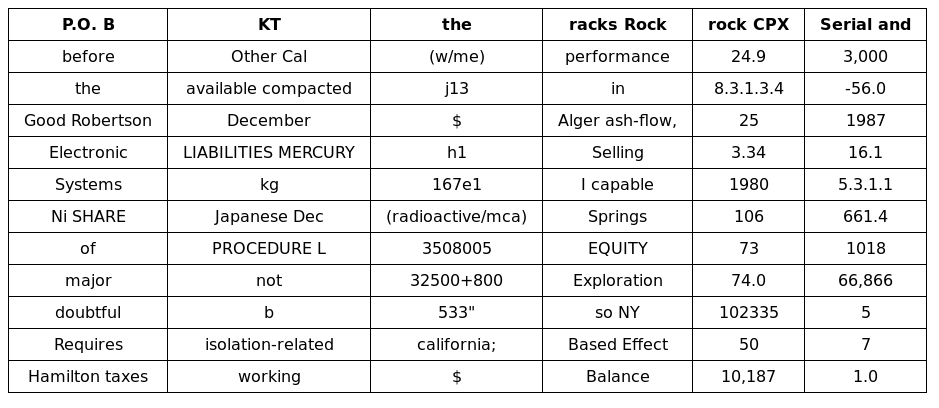}\label{fig:category1}}\quad
   \subfigure[Category 2]{\includegraphics[width=.35\textwidth]{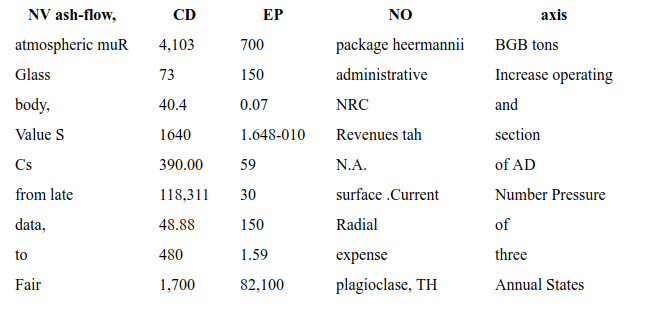}\label{fig:category2}}\\
   \subfigure[Category 3]{\includegraphics[width=.35\textwidth]{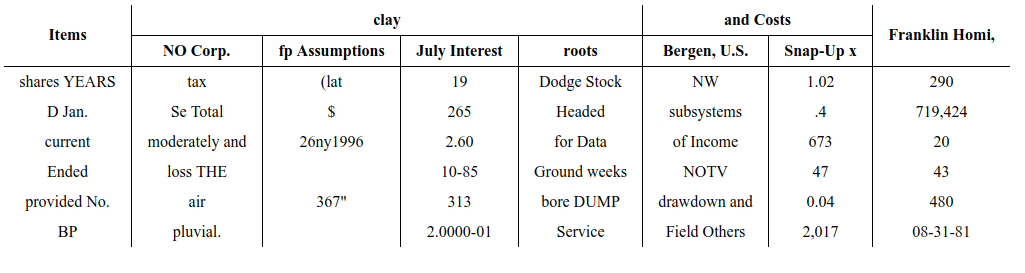}\label{fig:category3}}\quad
   \subfigure[Category 4]{\includegraphics[width=.35\textwidth]{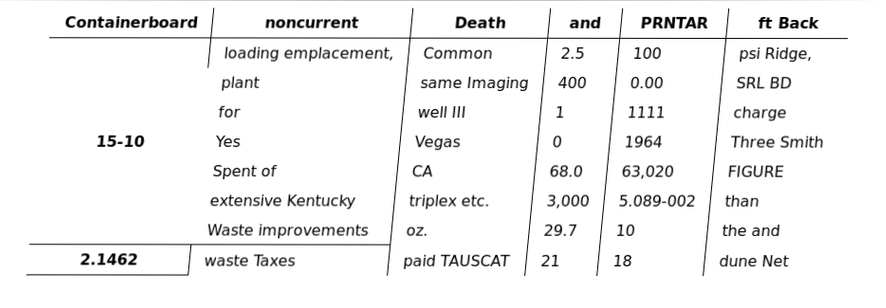}\label{fig:category4}}
   \caption{Images with different difficulty categories.  Category 1 images are plain images with no merging and with ruling lines. Category 2 adds different border types including occasional absence of ruling lines. Category 3 is the hardest one which introduces cell and column merging. Category 4 models camera captured images by linear perspective transform.}
   \label{fig:syntheticdataset}
\end{figure*}

\section{Introduction}
Structural analysis is one of the most important aspects of document processing. It incorporates both physical and logical layout analysis and also includes parsing or recognition of complex structured layouts including tables, receipts, and forms. While there has been a lot of research done in physical and logical layout analysis of documents, there is still ample room for contribution towards parsing of structured layouts, such as tables, within them. Tables provide an intuitive and natural way to present data in a format which could be readily interpreted by humans. Based on its significance and difficulty level, table structure analysis has attracted a large number of researchers to make contributions in this domain.

Table detection and recognition is an old problem with research starting from the late nineties. One of the initial work is by Kieninger et al.~\cite{trecs}. They used a bottom-up approach on words bounding boxes using a heuristics-based  algorithm. Later on, many different hand-crafted features based methods were introduced including~\cite{wangt2001automatic},~\cite{hu1999medium},~\cite{gatos2005automatic} and~\cite{tupaj1996extracting} which relied on custom-designed algorithms. Zanibbi et al.~\cite{zanibbi2004survey} present a comprehensive survey of table detection and structure recognition algorithms at that time. An approach to recognize tables in spreadsheets was presented by~\cite{doush2010detecting} which classified every cell into either a header, a title, or a data cell. Significant work was done by Shafait et al.~\cite{shafait2010table} where they introduced different performance metrics for the table detection problem. These approaches are not data driven and they make strong assumptions about tabular structures.

Chen et al.~\cite{chenin11} used support vector machines and dynamic programming for table detection in handwritten documents. Kasar et al.~\cite{kasar2013learning} also used SVMs on ruling lines to detect tables. Hao et al.~\cite{hao2016table} used loose rules for extracting table regions and classified the regions using CNNs. They also used textual information from PDFs to improve the model results. Rashid et al.~\cite{sheikh17} used positional information in every word to classify it as either a table or a non-table using dense neural networks.

After 2016, the research trod towards using deep learning models to solve the challenge. In 2017, many papers were presented which used object detection or segmentation models for table detection and parsing. Gilani et al.~\cite{gilani2017table} employed distance transform encoded information in an image and applied Faster RCNN~\cite{girshick2015fast} on these images. Schreiber et al.~\cite{schreiber2017deepdesrt} also used Faster RCNN for table detection and extraction of rows and columns. For parsing, they applied object detection algorithm on vertically stretched document images. Leveraging the tables' property to empirically contain more numeric data than textual data, Arif et al.~\cite{forebackfeatures} proposed to color code the document image to distinguish the numeric text and applied faster RCNN to extract table regions.
Similarly, Siddique et al.~\cite{decnt} presented an end-to-end Faster-RCNN pipeline for table detection task and used Deformable Convolutional Neural Network~\cite{deformablecnn} as feature extractor for its capability to mold its receptive field based on the input.
He et al.~\cite{fcndefang} segmented the document image into three classes: text, tables, and figures. They proposed to use Conditional Random Field (CRF) to improve results from Fully Convolutional Network (FCN) conditioned on the output from the contour edge detection network. Kavasidis et al.~\cite{saliencybased} employed CRFs on saliency maps extracted by Fully Convolutional Neural Network to detect tables, and different type of charts.

\begin{figure*}[h!]
   \centering
   \includegraphics[width=.95\textwidth]{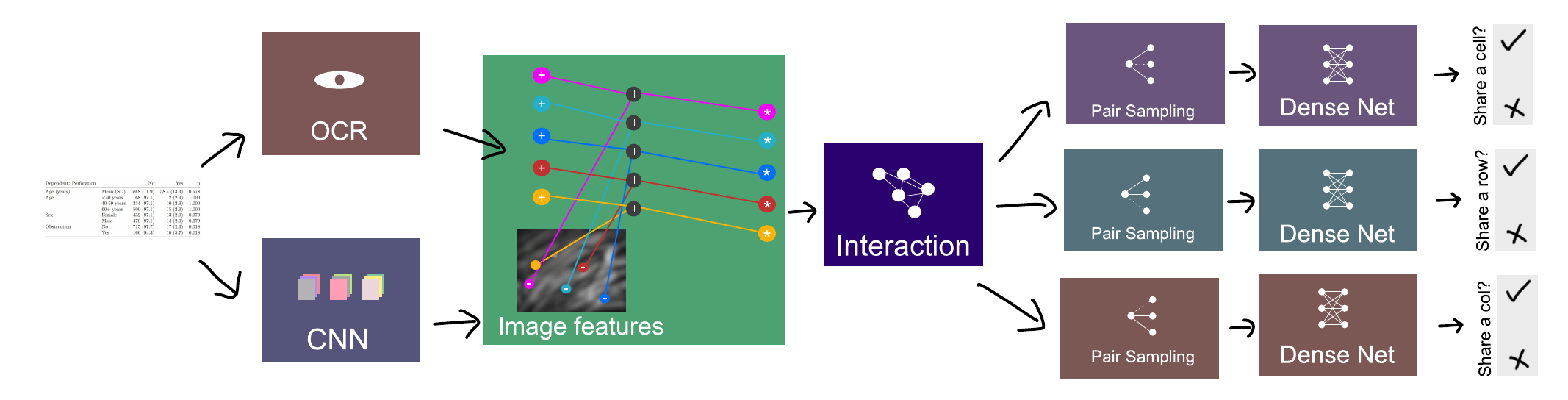}
   \caption{A pictorial representation of our architecture. For a document image, a feature map is extracted using the CNN model and words' positions are extracted using an OCR engine or an oracle (+). Image features, corresponding to the words' positions, are gathered (-) and concatenated ($||$) with positional features to form the input features (*). An interaction network is applied to the input features to get representative features. For each vertex (word), sampling is done individually for cells, rows and columns classification. The representative features for every sample pair are concatenated again and used as the input for the dense nets. These dense nets are different for rows, columns and cells sharing. Sampling is only done during the training. During the inference, every vertex pair is classified.}
    \label{fig:architecture}
   
\end{figure*}

Even though many researchers have shown that using object detection based approaches work well for table detection and recognition, defining the parsing problem in the form of object-detection problem is hard, especially if the documents are camera captured and contain perspective distortions. Approaches like~\cite{schreiber2017deepdesrt} partially solve the issue but it is still not a natural approach. It also makes it harder to use further features which could be extracted independently, for instance, the language features which could possibly hint towards the existence of a table.


In this paper, we define the problem using graph theory and apply graph neural networks to it. One of the initial research in graph neural networks is done by Scarselli et al.~\cite{scarselli2009graph} where they formulated a comprehensive graph model based on contraction maps. In recent years, they have gained a lot of traction due to the increase in compute power and with the introduction of newer methods. Many notable works include \cite{gilmer2017neural}, \cite{defferrard2016convolutional}, \cite{li2015gated}. Battaglia et al.~\cite{battaglia2018relational} argued that relational inductive biases are the key to achieving human like-intelligence and showed how graph neural networks are essential for it.

The use of graphs in document processing is not new. There have been many papers published which employ graph-based models for a wide range of problems. Liang et al.~\cite{liang2001optimization} introduced a hierarchical tree-like structure for document parsing. Work presented by Wang~\cite{wang2016tabular} gained a lot of popularity and was used by many researchers afterward. Hu et al.~\cite{hu2000table} introduced a comprehensive graph model involving a Directed Acyclic Graph (DAG) with detailed definitions of various elements of a table. Recently, Koci et al.~\cite{koci2018table} presented an approach where they encoded information in the form of a graph. Afterward, they used a newly-proposed rule-based remove-and-conquer algorithm. Bunke et al.~\cite{BunkeBunke} provides a detailed analysis of different graph-based techniques employed in the context of document analysis. These methods make strong assumptions about the underlying structure which contradicts with the philosophy of deep learning. Even though we are not the first ones to use graphs for document processing, to the best of our knowledge, we are the first ones to apply graph neural networks to our problem. We have done our experiments on the table recognition problem, however, this new problem definition applies to various other problems in document structural analysis. There are two advantages to our approach. For one, it is more generic since it doesn't make any strong assumptions about the structure and it is close to how humans interpret tables, i.e. by matching data cells to their headers. Secondly, it allows us to exploit graph neural networks towards which there has been a lot of push lately.

In particular, we make the following contributions: 
\begin{enumerate}
    \item Formulate table recognition problem as a graph problem which is compatible with graph neural networks
    \item Design a novel differentiable architecture which reaps the benefits of both convolutional neural networks for image feature extraction and graph neural networks for efficient interaction between the vertices
\item Introduce a novel Monte Carlo based technique to reduce memory requirements of training
    \item Fill the gap of large scale dataset by introducing a synthetic dataset
    \item Run tests on two state-of-the-art graph based methods and emperically show that they perform better than a baseline network
\end{enumerate}

\section{Dataset}
There are a few datasets for table detection and structure recognition published by the research community, including UW3, UNLV~\cite{unlvuw3} and ICDAR 2013 table competition dataset~\cite{gobel2013icdar}. However, the size of all of these datasets is limited. It risks overfitting in deep neural networks and hence, poor generalization. Many people have tried techniques such as transfer learning but these techniques cannot completely offset the utility of a large scale dataset.

We present a large synthetically generated dataset of 0.5 Million tables divided into four categories, which are visualized in Figure~\ref{fig:syntheticdataset}. To generate the dataset, we have employed Firefox and Selenium to render synthetically generated HTML. We note that synthetic dataset generation is not new and similar work~\cite{wangt2001automatic} has been done before. Even though it will be hard to generalize algorithms from our dataset to the real world, the dataset provides a standard benchmark for studying different algorithms until a large scale real-world dataset is created. We have also published our code to generate further data, if required.

\section{The graph model}
\label{sec:graphmodel}
Considering the problem of table recognition, the ground truth is defined as three graphs wherein every word is a vertex. There are three adjacency matrices representing each of the graphs, namely cell-, row-, and column-sharing matrices. So if two vertices share a row i.e. both words belong to the same row, these vertices are taken to be adjacent to each other (likewise for cell and column sharing).

\begin{figure}[ht]
\centering
\includegraphics[width=0.8\linewidth]{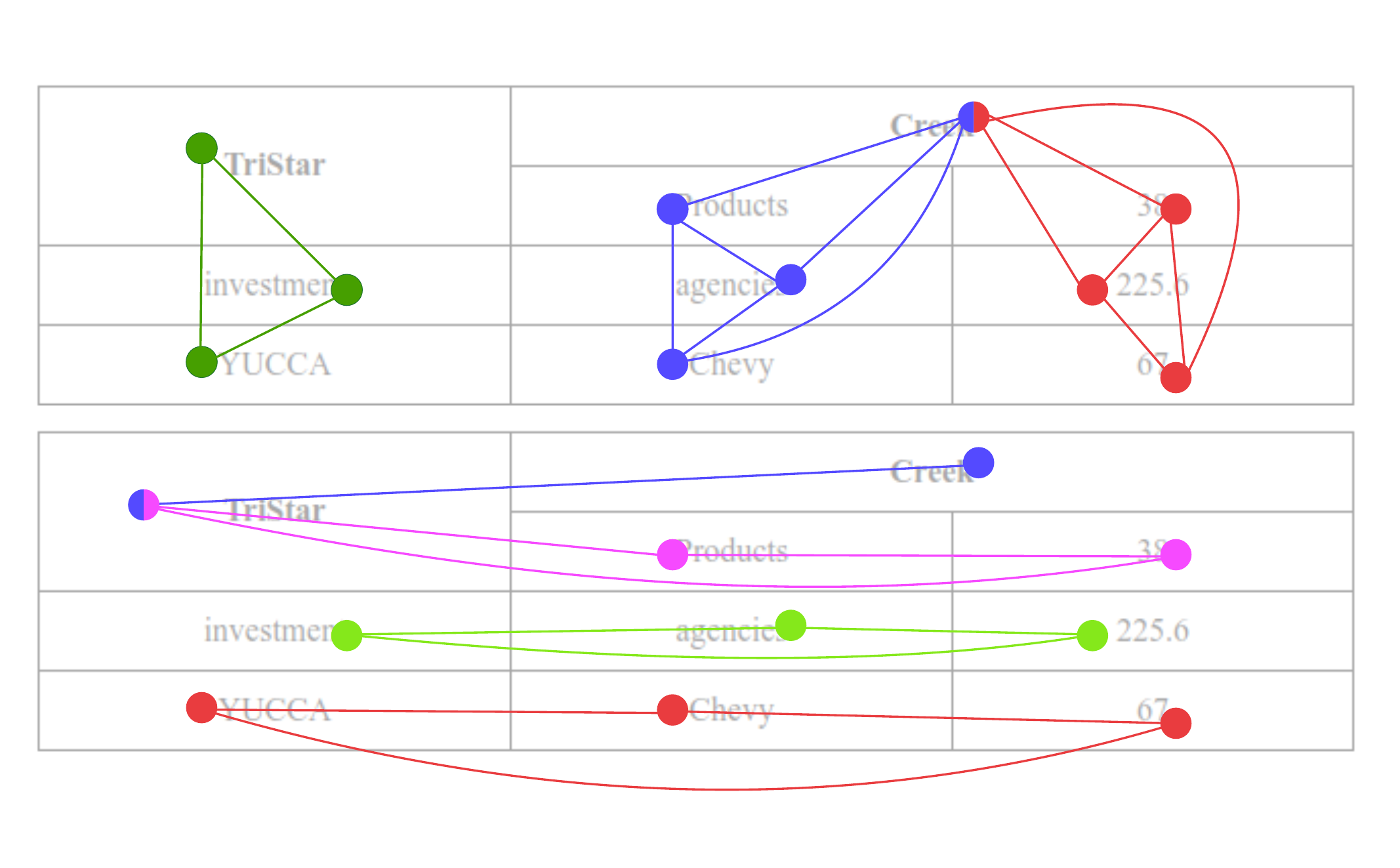}
\caption{Reconstructing the resultant column and row segments using maximal cliques. The upper figure shows column cliques while the bottom one shows row cliques. Note the merged vertex in both of these figures which belongs to multiple cliques.}
\label{fig:clique}
\end{figure}

The prediction of a deep model is also done in the form of the three adjacency matrices. After getting adjacency matrices, complete cells, rows and columns can be reconstructed by solving the problem of maximal cliques~\cite{bron1973algorithm} for rows and columns and connected components for cells. It is pictorially shown in Figure~\ref{fig:clique}.

This model is valid not only for table recognition problem but can also be used for document segmentation. In that scenario, if two vertices (could be words again) share the same zone, they are adjacent. The resultant zones can also be reconstructed using the maximal clique problem.

\section{Methodology}
All of the tested models follow the same parent pattern, shown in Figure~\ref{fig:architecture}, divided into three parts: the convolutional neural network for extraction of image features, the interaction network for communication between the vertices, and the classification part to label every paired vertices as being adjacent or not adjacent (class 0 or class 1) in each of the three graphs.

The algorithm for the forward pass is also given in Algorithm~\ref{alg:forward}. It takes the image ($I \in \mathbb{R}^{h \times w \times c}$ --- where $h$, $w$ and $c$ represent height, width and number of channels in the input image respectively), positional features ($F_p \in \mathbb{R}^{v \times 4}$ --- where $v$ represents number of vertices), and other features $F_o\in \mathbb{R}^{v \times o}$. In addition to this, it also takes the number of samples per vertex($s\in \mathbb{R}$) and three adjacency matrices ($A_{cells}\in \{0,1\}^{v \times v}$, $A_{rows}\in \{0,1\}^{v \times v}$ and $A_{cols}\in \{0,1\}^{v \times v}$) as the input during the training process. All the parametric functions are denoted by $f$ and non-parametric functions by $g$. If all the parametric functions are differentiable, the complete architecture will be differentiable as well and hence, compatible with backpropagation.

The positional features include the coordinates of the upper left and the bottom right corner of each vertex. Other features consist only of the length of the word in our case. However, in a real-world dataset, natural language features~\cite{pennington2014glove} could also be appended which may provide additional information.

\subsubsection{Convolutional neural network}
A convolutional neural network ($f_{conv}$) takes an image ($I$) as its input and as the output, it generates the respective convolutional features ($I_f \in \mathbb{R}^{h' \times w' \times q}$ --- $w'$, $h'$ and $q$ being the width, height and number of channels of the convolutional feature map respectively). To keep parameter count low, we have designed a shallow CNN; however, any standard architecture can be used in its place. At the output of CNN, a gather operation ({$g_{gather}$}) is performed to collect convolutional features for each word corresponding to its spatial position in the image and form gathered features($F_{im}\in \mathbb{R}^{v \times q}$). Since convolutional neural networks are translation equivariant, this operation works well. If the spatial dimensions of the output features are not the same as the input image (for instance, in our case, they were scaled down), the collect positions are linearly scaled down depending on the ratio between the input and output dimensions. The convolutional features are extended to the rest of the vertex features ($g_{ext}$).

\subsubsection{Interaction}
After gathering all the vertex features, they are passed as input to the interaction model ($f_{int}$). We have tested two graph neural networks to use as the interaction part which are the modified versions of~\cite{wang2018dynamic} and~\cite{caloGraphNN} respectively. These modified networks are referred to as DGCNN* and GravNet* hereafter. In addition to these two, we have also tested with a baseline dense net (dubbed FCNN for Fully Connected Neural Network) with approximately the same number of parameters to show that the graph-based models perform better. For these three models, we have limited the total parameter count to $1M$ for a fair comparison. This parameter count also includes parameters of preceding CNN and the succeeding classification dense network. As the output, we get representative features ($F_{rep}\in \mathbb{R}^{v \times r}$ --- $r$ being the number of representative features) of each of the vertex which are used for classification.

\subsubsection{Runtime pair sampling}
Classifying every word pair is a memory intensive operation with memory complexity of ${O(N^2)}$. Since it would then scale linearly with the batch size, the memory requirements increase even further. To cater to this, we employed a Monte Carlo based sampling. The index sampling function is denoted by $g_{sample}$. this function would generate a fixed number of samples ($s$) for each of the vertex for each of the three problems (cell sharing, row sharing and column sharing).

Uniform sampling is highly biased towards class 0. Since we can't use a large batch size due to the memory constraints, the statistics are not sufficient to differentiate between the two classes. To deal with this issue, we changed the sampling distribution ($P_{samples}$) to sample, on average, an equal number of elements of class 0 and class 1 for each of the vertex. It can be easily done in a vectorized fashion as shown in Algorithm~\ref{alg:forward}. Note that $J$ in the algorithm denotes an all-one matrix. Different sets of samples are collected for each of the three classes for each of the vertex ($S_{cells}\in \mathbb{Z}^{v \times t}$, $S_{rows}\in \mathbb{Z}^{v \times t}$, $S_{cols}\in \mathbb{Z}^{v \times t}$). The values in these matrices represent the index of the paired samples for each of the vertex. For inference, however, we do not need to sample since we don't need to use the mini-batch approach. Hence, we simply do it for every vertex pair. So, during training, $t=s$ and during inference, $t=v$.

\begin{algorithm}[h!]
    \caption{Forward Pass}\label{alg:forward}
    \hspace*{\algorithmicindent} \textbf{Input} $I,F_p,F_o, (A_{cells}, A_{rows}, A_{cols}, s)\footnotemark$\\
    \hspace*{\algorithmicindent} \textbf{Output}$L_{cells}, L_{rows}, L_{cols}$
    \begin{algorithmic}[1]
    \Function{Pair\_Sampling}{$A$, $s$}
      \State $P_{samples} \gets 0.5*(1-A) \cdot ((1-A) \times J_{v,v})^{\circ-1}$
      \State $P_{samples} \gets P_{samples} + 0.5*A \cdot (A \times J_{v,v})^{\circ-1}$
      \State \Return $g_{sample}(P_{samples}, s)$
    \EndFunction
    
    \State $I_f \gets f_{conv}(I)$
    \State $F_{im} \gets g_{gather}(I_f, F_p)$
    \State $F_{cat} \gets g_{concat}(F_{im}, F_o, F_p)$
    \State $F_{int} \gets f_{int}(F_{cat})$
    \If {training}
        \State $S_{cells} \gets \textproc{Pair\_Sampling}(A_{cells}, s)$
        \State $S_{rows} \gets \textproc{Pair\_Sampling}(A_{rows}, s)$
        \State $S_{cols} \gets \textproc{Pair\_Sampling}(A_{cols}, s)$
    \Else
    \State $S_{cells} \gets S_{rows} \gets S_{cols} \gets [x_{i,j}]_{v,v} \mid x_{i,j}=j$
    \EndIf
    
    \State $L_{cells} \gets f_{cells}(S_{cells}, F_{int})$
    \State $L_{rows} \gets f_{rows}(S_{rows}, F_{int})$
    \State $L_{cols} \gets f_{cols}(S_{cols}, F_{int})$
    
    \end{algorithmic}
\end{algorithm}
\footnotetext{Only needed for training}

\subsubsection{Classification}
After sampling, the elements from the output feature vector ({$F_{int}$}) of the interaction model and the elements from the sampling matrices are concatenated ($g_{cat}$) with each other in ($f_{cells}$, $f_{rows}$, and $f_{cols}$). These functions are parametric neural networks. As the output, we get three sets of logits $L_{cells} \in \mathbb{R}^{v \times t \times 2}$, $L_{rows} \in \mathbb{R}^{v \times t \times 2}$ and $L_{cols} \in \mathbb{R}^{v \times t \times 2}$. They can be used either to compute the loss and backpropagate through the function, or to predict the classes and form the resultant adjacency matrices.

\begin{table*}[t]
    \centering
    \caption{True positive rate: the percentage of the GT cells/rows and the column cliques in category $i$ which have a match in the predictions.}
    \begin{tabular}{l|c|c|c|c|c|c|c|c|c|c|c|c}
          Method & $cells_1$ & $rows_1$ & $cols_1$ &$cells_2$ & $rows_2$ & $cols_2$ &$cells_3$ & $rows_3$ & $cols_3$ &$cells_4$ & $rows_4$ & $cols_4 $  \\ \hline
          FCNN & \textbf{99.9} & 99.9 & 99.6 & \textbf{99.9} & 99.6 & 99.4 & 99.8 & 96.8 & 87.6 & \textbf{99.9} & 97.7 & 90.0 \\
          GravNet* & 99.8 & \textbf{100} & 99.7 & 99.8 & \textbf{99.9} & 99.5 & 99.2 & 95.7 & 86.2 & 99.6 & 96.8 & 90.5 \\
          DGCNN* & 99.8 & 99.9 & \textbf{100} & \textbf{99.9} & \textbf{99.9} & \textbf{99.8} & \textbf{99.9} & \textbf{98.1} & \textbf{94.1} & 99.8 & \textbf{99.1} & \textbf{94.3} \\
    \end{tabular}
   
    \label{tab:gtinpred}
\end{table*}

\begin{table*}[t]
    \centering
    \caption{False positive rate: the percentage of the detected cells/rows and the column cliques in category $i$ which do not have a match in the ground truth.}
    \begin{tabular}{l|c|c|c|c|c|c|c|c|c|c|c|c}
          Method & $cells_1$ & $rows_1$ & $cols_1$ &$cells_2$ & $rows_2$ & $cols_2$ &$cells_3$ & $rows_3$ & $cols_3$ &$cells_4$ & $rows_4$ & $cols_4 $  \\ \hline
              FCNN & \textbf{0.01} & 0.5 & 16.7 & \textbf{0.06} & 3.05 & 14.1 & 0.12 & 10.4 & 32.1 & \textbf{0.04} & 6.31 & 26.4 \\
          GravNet* & 0.15 & \textbf{0.18} & 6.56 & 0.17 & 0.79 & 9.01 & 0.58 & 10.2 & 33.6 & 0.28 & 7.81 & 25.28 \\
          DGCNN* & 0.07 & 0.46 & \textbf{0.79} & 0.08 & \textbf{0.22} & \textbf{1.09} & \textbf{0.06} & \textbf{4.8} & \textbf{14.6} & 0.07 & \textbf{4.39} & \textbf{10.6} \\
    \end{tabular}
   
    \label{tab:prednotgt}
\end{table*}

\begin{table*}[t]
    \centering
    \caption{End-to-end accuracy/perfect matching.}
    \begin{tabular}{l|c|c|c|c}
          Method & Category $1$ & Category $2$ & Category $3$ & Category $4$  \\ \hline
          FCNN & 42.4 & 54.6 & 10.9 & 31.9 \\
          GravNet* & 65.6 & 58.6 & 13.1 & 31.5 \\
          DGCNN* & \textbf{96.9} & \textbf{94.7} & \textbf{52.9} & \textbf{68.5} \\
    \end{tabular}
    
    \label{tab:perf}
\end{table*}

\begin{figure}[b!]
\captionsetup[subfigure]{justification=centering}
    \centering

      \subfigure[Cells Predictions]{
      \includegraphics[width=8cm, height=2.0cm]{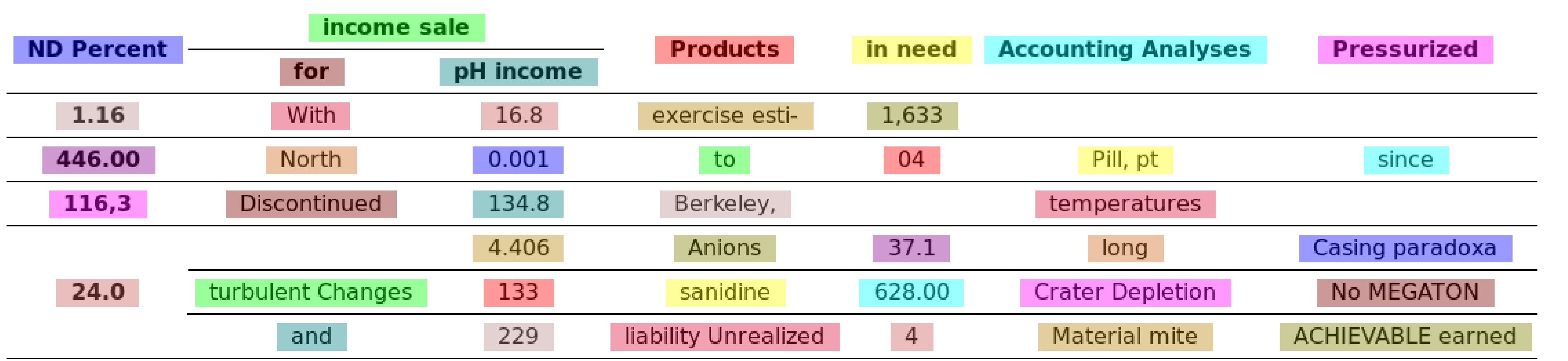}
      \label{fig:cellspred}
      }
      
      \vspace{2mm}
      
      \subfigure[Rows Predictions]{
      \includegraphics[width=8cm, height=2.0cm]{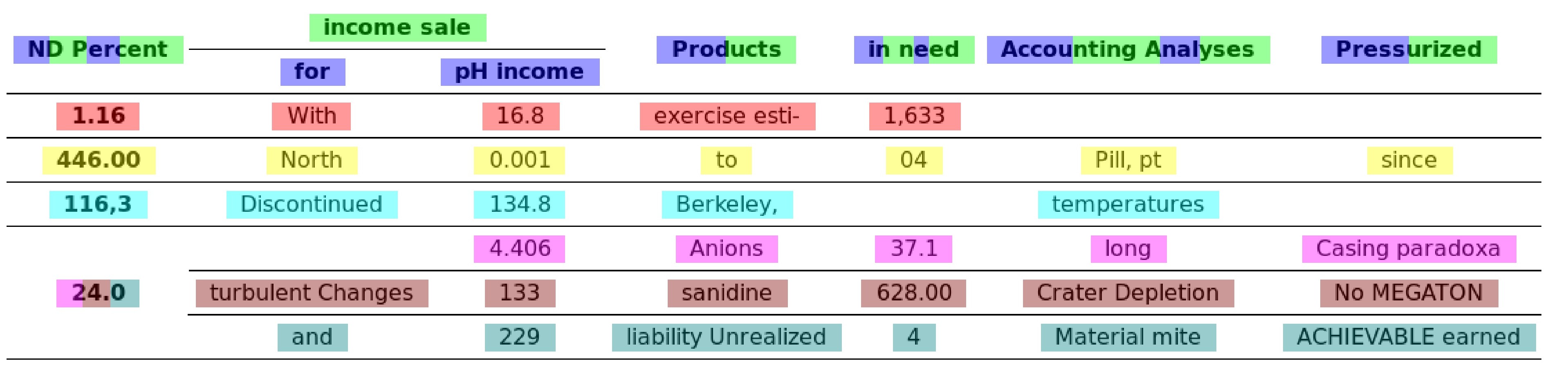}
      \label{fig:rowspred}
      }
      
      \vspace{2mm}
      
      \subfigure[Columns Predictions]{
      \includegraphics[width=8cm, height=2.0cm]{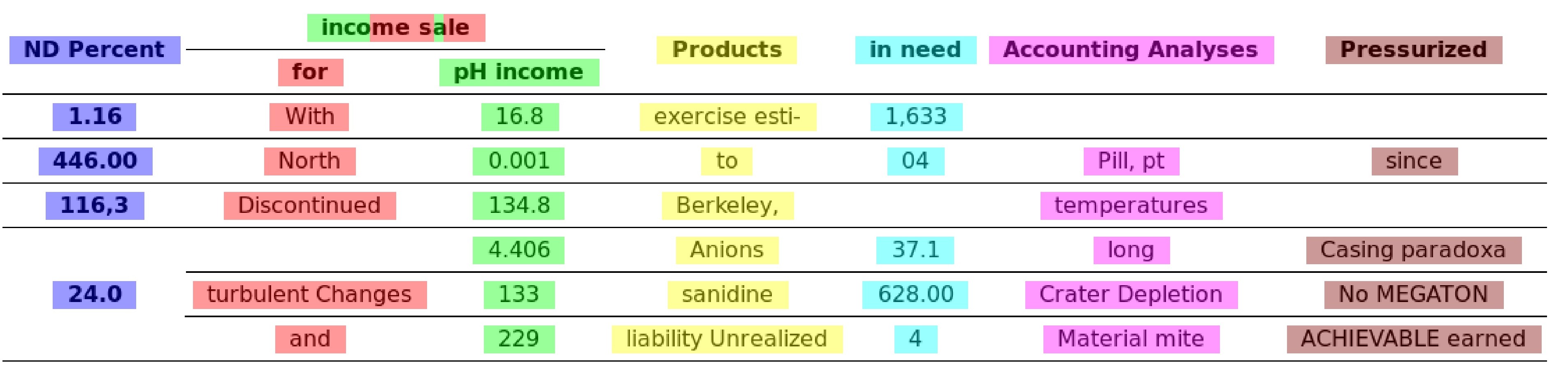}
      \label{fig:colspred}
      }
      
      \caption{Figures show DGCNN* model's predictions (cells, rows and columns). \ref{fig:cellspred} shows the cells predictions. All the words in a single cell are assigned the same color. \ref{fig:rowspred} shows the rows predictions. The words belonging to the same row are assigned the same color and the words spanned across multiple rows (cliques) are stripe-colored. Similarly, \ref{fig:colspred} shows the column predictions.}
      \label{fig:resultimages2}
\end{figure}

\begin{figure}[h!]
\captionsetup[subfigure]{justification=centering}
    \centering
    
      \subfigure[Adjacent in Cells]{
      \includegraphics[width=8cm,keepaspectratio]{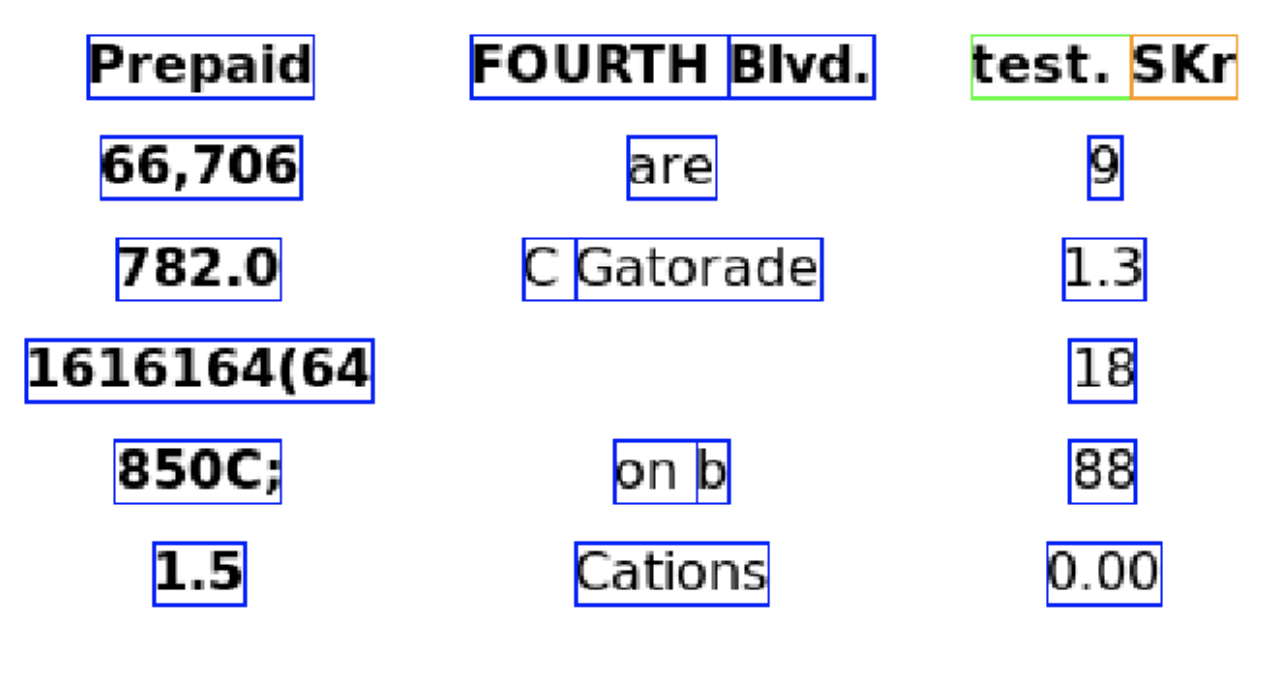}
      \label{fig:cellvis}
      }
      
      \vspace{2mm}
      
      \subfigure[Adjacent in Rows]{
      \includegraphics[width=8cm,keepaspectratio]{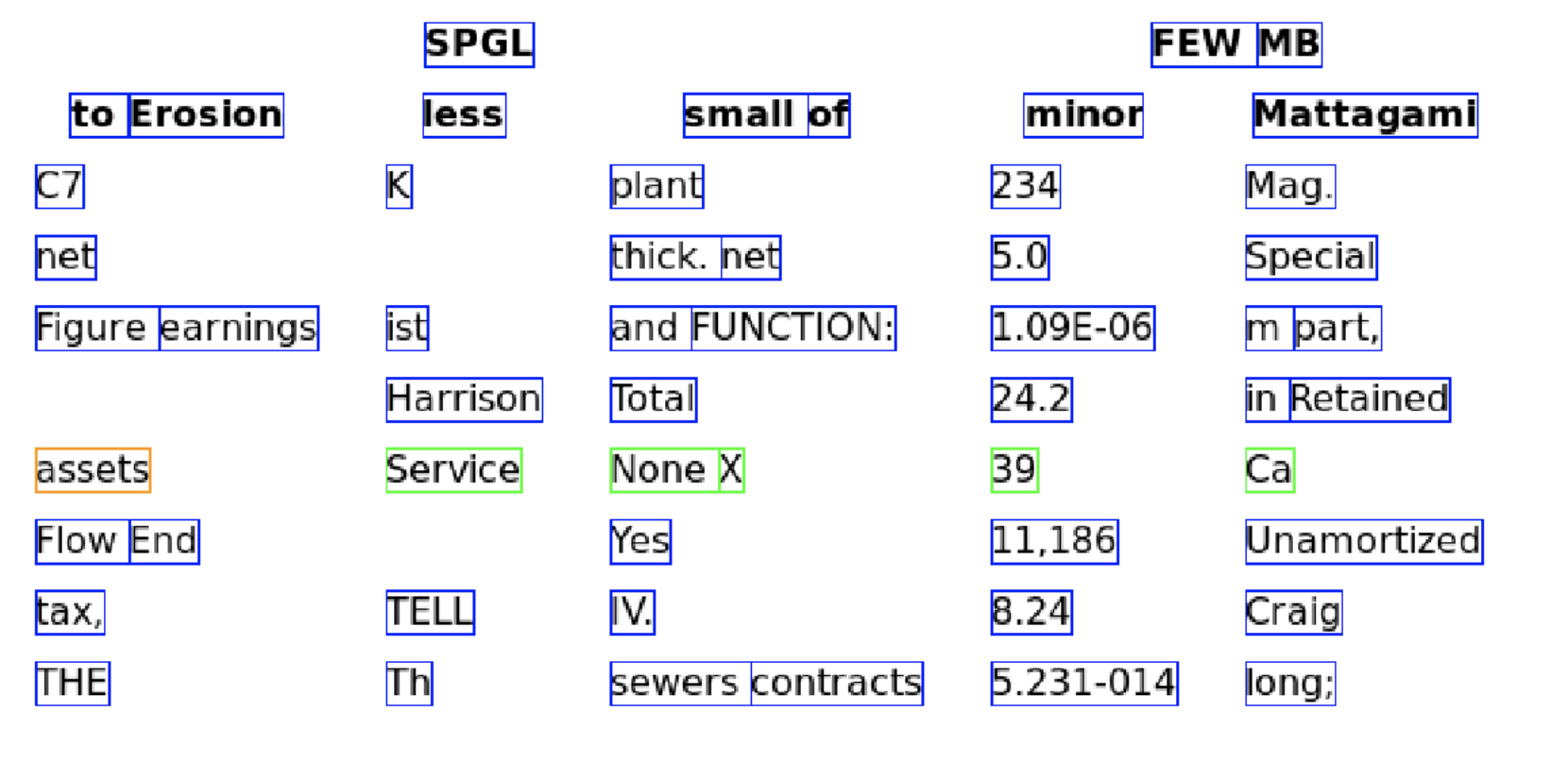}
      \label{fig:rowvis}
      }
      
      \vspace{2mm}
      
      \subfigure[Adjacent in Columns]{
      \includegraphics[width=8cm,keepaspectratio]{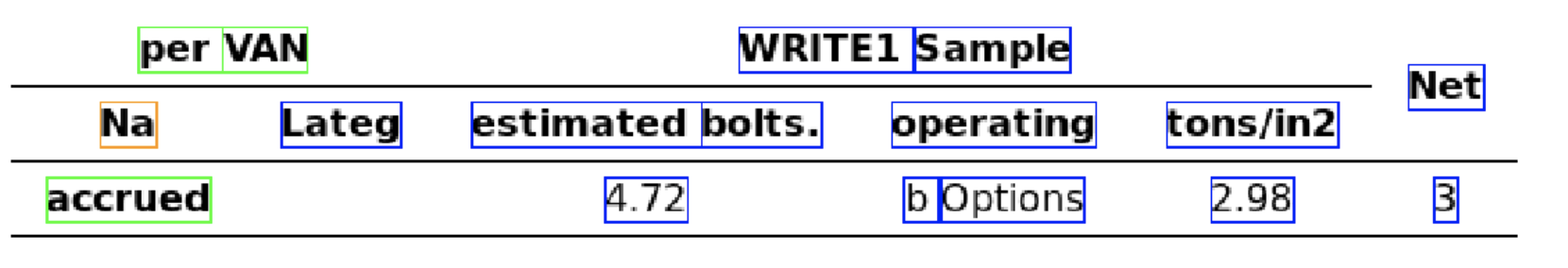}
      \label{fig:colvis}
      }
      
      \caption{Figures show DGCNN* model's predictions (cells, rows and columns neighbors) for a randomly selected vertex shown in orange. \ref{fig:cellvis} shows the cells predictions. All the vertices adjacent to the random vertex in cell-sharing graph are shown in green and those not adjacent are shown in blue. Likewise \ref{fig:rowvis} and \ref{fig:colvis} show the adjacent neighbors in row- and column sharing graphs respectively.}
      \label{fig:resultimages}
\end{figure}

\section{Results}
Shahab et al \cite{shahab2010open} defined a set of metrics for detailed evaluation of the results of table parsing and detection. They defined criteria for correct and partial detection and defined a heuristic for labeling elements as under-segmented, over-segmented and missed. Among their criterion, two are the most relevant to our case, i.e. the percentage of the ground truth elements that are detected correctly (true positive rate) and the number of the predicted elements which do not have a match in the ground truth (false positive rate). In our case, as argued in Section~\ref{sec:graphmodel}, the elements are cliques. So true positive rate and false positive rate is computed on all three graphs (cells, rows and columns) individually. This rate is averaged over the whole test set. These results are shown in Table~\ref{tab:gtinpred} and Table~\ref{tab:prednotgt}.

In addition to this, we also introduce another measure, i.e. perfect matching as shown in Table~\ref{tab:perf}. If all of the three predicted adjacency matrices are perfectly matched with the corresponding matrices in the ground truth, the parsed table is labeled as being end-to-end accurate. This is a strict metric but it shows how misleading can the statistics computed on the basis of individual table elements be due to the large class imbalance problem.

As expected, since there is no interaction between vertices in the FCNN, it performs worse than the graph models. Note however, that the vertices in this network are not completely segregated. They can still communicate in the convolutional neural network part. This builds the case to introduce graph neural networks further into document analysis.

Category 3 tables show relatively poor results as compared to category 4 tables. This is because category 4 images also contain those from category 1 and 2 to study the effect of perspective distortion on simpler images. We conclude that while the graph networks struggle with merged row and columns, they gracefully handle the perspective distortions.

\section{Conclusion and future work}
In this work, we redefined the structural analysis problem using the graph model. We demonstrated our results on the problem of table recognition and we also argued how several other document analysis problems can be defined using this model. Convolutional neural networks are the most suited at finding representative image features and graph networks are the most suited at fast message passing between vertices. We have shown how we can combine these two abilities using the gather operation. So far, we only used positional features for the vertices, but for a real-world dataset, natural language processing features like GloVe can also be used. In conclusion, graph neural networks work well for structural analysis problems and we expect to see more research in this direction in the near future.

\bibliographystyle{IEEEtran}
\bibliography{bare_conf}

\end{document}